\documentclass[11pt]{article}
\usepackage[preprint]{acl}
\usepackage{amsmath}
\usepackage{times}
\usepackage{latexsym}
\usepackage[T1]{fontenc}
\usepackage[utf8]{inputenc}
\usepackage{microtype}
\usepackage{graphicx}
\usepackage{booktabs}
\newcounter{appendixalgorithm}[section]
\renewcommand{\theappendixalgorithm}{\Alph{section}.\arabic{appendixalgorithm}}
\newcounter{pseudostep}
\newenvironment{pseudocode}[2]{%
  \par\addvspace{0.7\baselineskip}\noindent
  \begin{minipage}{\linewidth}
  \refstepcounter{appendixalgorithm}\label{#2}
  \hrule height 0.6pt\vspace{4pt}
  {\small\textbf{Algorithm \theappendixalgorithm}\quad #1\par}
  \vspace{4pt}\hrule height 0.3pt\vspace{3pt}
  \small
  \begin{list}{\arabic{pseudostep}.}{%
    \usecounter{pseudostep}\setlength{\leftmargin}{1.6em}%
    \setlength{\labelwidth}{1.2em}\setlength{\labelsep}{0.4em}%
    \setlength{\rightmargin}{0pt}\setlength{\itemsep}{2pt}%
    \setlength{\parsep}{0pt}\setlength{\topsep}{3pt}}
}{%
  \end{list}\vspace{3pt}\hrule height 0.6pt
  \end{minipage}\par\addvspace{0.7\baselineskip}}
\newcommand{\pstep}[2][0]{\item\hspace*{#1em}#2}
\makeatletter
\newenvironment{appendixtable}{%
  \par\addvspace{0.7\baselineskip}\noindent
  \begin{minipage}{\linewidth}\captionsetup{type=table,hypcap=false}
}{%
  \end{minipage}\par\addvspace{0.7\baselineskip}}
\makeatother
\newcommand{\pinput}[1]{\item[]\textbf{Input:} #1}
\newcommand{\poutput}[1]{\item[]\textbf{Output:} #1}

\title{{ReVision}: Selective Intrinsic Reasoning
for Tool-Augmented, Reasoning-Intensive VQA}
\author{%
  Ningxin Pan\thanks{Equal contribution.}\quad
  Guochen Yan\footnotemark[1]\quad
  Jialong Wu\quad
  Wentao Zhang\thanks{Corresponding author: \href{mailto:wentao.zhang@pku.edu.cn}{wentao.zhang@pku.edu.cn}.}\\
  Peking University
}

\begin{document}

\maketitle

\begin{abstract}
External tools substantially expand what multimodal large language models (MLLMs) can retrieve and compute on reasoning-intensive visual questions, and aggregate accuracy typically improves when tools are available. Our empirical study across six VQA benchmarks shows that aggregate accuracy gains under tool access can coexist with \emph{answer reversals}: questions answered correctly without tools but incorrectly with tools. Trajectory analysis reveals substantially shorter reasoning turns under tool access, suggesting a shift from intrinsic visual reasoning toward reliance on external outputs. We propose \textsc{ReVision}, a training-free test-time method that registers intrinsic reconsideration as an optional action alongside search and code execution. When invoked, a separate tool-free multimodal call re-examines the original image, integrates accumulated evidence, and returns a revised analysis. The MLLM chooses if and when to use it, without fixed reflection stages or paired rollouts. \textsc{ReVision} improves over standard tool use on all six benchmarks (1.28--7.51\%) and outperforms Dual-Rollout Combine on five, with reasoning-token use comparable to standard tool use.
\end{abstract}

\section{Introduction}

Multimodal large language models (MLLMs) combine visual understanding with language-based reasoning to answer questions about images. In scientific and knowledge-intensive applications, these questions often require specialized knowledge or calculations beyond visual recognition. Tool-augmented MLLMs address these demands by integrating external search and code execution into reasoning~\citep{yang2023mmreact,lu2023chameleon,liu2025llavaplus}. They therefore provide an important setting for studying how models coordinate visual reasoning with external evidence.

However, expanding an MLLM's external capabilities does not guarantee that it will use all available evidence effectively. A tool-augmented model must still determine how retrieved information relates to the original image and whether it supports the model's current interpretation. Our empirical study across six VQA benchmarks shows that aggregate accuracy gains under tool access can coexist with \emph{answer reversals}: questions answered correctly without tools but incorrectly with tools. Across six reasoning-intensive VQA benchmarks and 11{,}572 paired questions, these reversals affect 9.70\% of all evaluated questions (Table~\ref{tab:answer-reversals}). Thus, although tools improve overall accuracy, they can also lead the model away from solutions it reaches successfully without them. This discrepancy motivates examining how the model's own visual reasoning changes when external tools \mbox{become available}.

\begin{figure*}[!t]
    \centering
    \includegraphics[width=\textwidth]{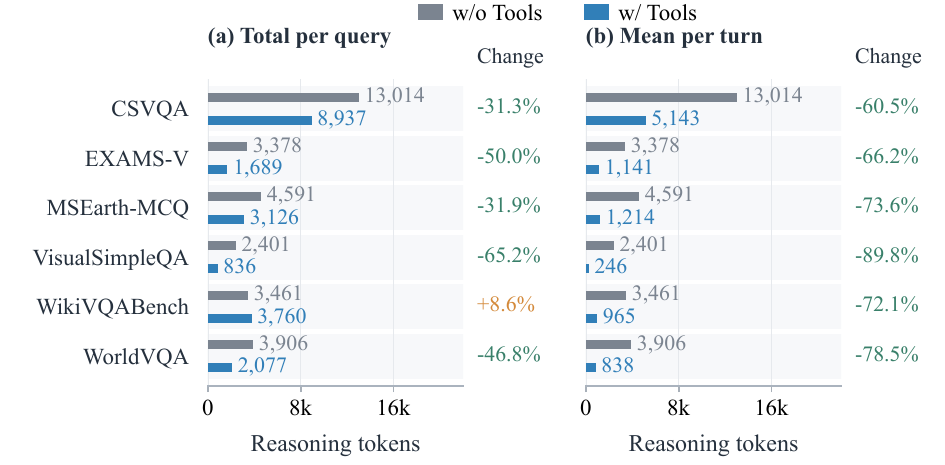}
    \caption{Reasoning-token comparison with and without external tools on six VQA benchmarks. Panels show (a) total tokens per query and (b) mean tokens per reasoning turn. Percentages indicate changes relative to w/o Tools.}
    \label{fig:compression}
\end{figure*}

To characterize this change, we examine how explicit reasoning changes under tool access. Reasoning turns become shorter on all six benchmarks, and total reasoning tokens per query decrease on five. This pattern suggests a shift toward external observations. Shorter reasoning alone does not establish why an answer becomes incorrect, but, together with answer reversals, it raises the concern that useful image-grounded reasoning may receive less attention as external observations accumulate. These findings motivate a mechanism for revisiting intrinsic visual reasoning and reconciling it with the evidence obtained through tools.

Combining tool-free and tool-augmented trajectories often improves accuracy, suggesting that the two modes retain complementary evidence. Our Dual-Rollout Combine baseline outperforms both individual inference modes on five of the six benchmarks. However, generating and fusing two complete trajectories incurs substantial reasoning-token cost, even when only a particular part of the agent's interpretation needs reconsideration. These findings lead to our research question: can a tool-using agent selectively revisit its intrinsic visual reasoning while benefiting from external evidence, within a single ongoing trajectory?

To address this challenge, we propose \textsc{ReVision}, a simple yet effective, training-free method that makes intrinsic multimodal reconsideration an optional action. Here, intrinsic reasoning refers to the MLLM's own reasoning over the image, question, and accumulated evidence without additional external tool calls. When invoked, a separate tool-free call to the same backbone re-examines the original image, reconciles accumulated evidence, and returns revised analysis to the ongoing trajectory. This action is available alongside \textsc{Search}, \textsc{Visit}, and \textsc{Python}. The MLLM decides whether and when to invoke this action, without a runner-enforced reflection schedule.

To evaluate \textsc{ReVision}, we conduct experiments on six reasoning-intensive VQA benchmarks. \textsc{ReVision} improves over standard tool use by 1.28--7.51\% and outperforms Dual-Rollout Combine on five benchmarks while using substantially fewer reasoning tokens. These gains are consistent with the view that selectively revisiting intrinsic visual reasoning helps the agent reconcile external evidence with the image.

Our contributions are threefold:
\begin{itemize}
    \item \textbf{Empirical Discovery:} We identify a tool-induced reasoning
    shift in multimodal VQA: enabling tools shortens explicit reasoning turns
    and produces answer reversals.
    \item \textbf{Method Design:} We propose \textsc{ReVision}, which treats
    intrinsic multimodal reconsideration as an optional callable action, letting
    the agent re-observe the image and reconcile evidence without leaving the
    trajectory.
    \item \textbf{Experimental Results:} \textsc{ReVision} improves over standard
    tool use on all six reasoning-intensive VQA benchmarks and exceeds
    Dual-Rollout Combine on five, while using substantially fewer reasoning
    tokens than Combine.
\end{itemize}

\section{Related Work}

\paragraph{Tool-Augmented Multimodal Reasoning.}
Tool-augmented agents extend model capabilities by combining reasoning with
external actions and observations, following ReAct~\citep{yao2023react}.
Multimodal systems add search, code execution, and visual perception~\citep{yang2023mmreact,lu2023chameleon,liu2025llavaplus}:
DeepEyes~\citep{zheng2026deepeyes} learns interleaved reasoning and active
visual inspection through reinforcement learning.
MMSearch-R1~\citep{wu2026mmsearchr1} learns on-demand image and text search.
ToolScope~\citep{deng2026toolscope} combines planning with iterative
Search, Code, and Perceive actions for long-horizon VQA. These systems raise
the question of how agents should integrate external observations with their
own reasoning.


\paragraph{Tool Reliance and Reasoning Offloading.}
\citet{guo2026multimodaltools} find limited consistent capability gains in
the multimodal agents they study: tool-call traces need not indicate
answer-critical information. \citet{lou2026toolillusion} show that visual
grounding remains beneficial in web agents with tools, although tool access
can reduce reliance on direct visual observations. The Tool-Overuse
Illusion~\citep{zeng2026tooloveruse} links unnecessary tool use to inaccurate assessments of the model's own knowledge and outcome-based incentives. The Theory of
Agent~\citep{wang2025toa} treats reasoning and external interaction as
complementary means of acquiring knowledge, advocating epistemically
necessary tool use. These findings motivate examining whether agents continue
using available visual evidence and intrinsic reasoning as external
capabilities expand.

\paragraph{Adaptive Tool Use and Reasoning Allocation.}
SMART~\citep{qian2025smart} uses supervised rationales to distinguish
steps solvable through intrinsic reasoning from tool-dependent ones. MeCo~\citep{li2025meco}
guides tool invocation through metacognitive signals from model
representations. AutoTool~\citep{ma2026autotool} uses reinforcement learning
to select between tool-assisted and text-centric reasoning according to the
query. Where these methods optimize tool-use decisions, \textsc{ReVision}
adds a training-free reconsideration action that the agent can invoke within
an ongoing tool-augmented trajectory, revisiting intrinsic reasoning as
external evidence accumulates.

\paragraph{Self-Reflection and Test-Time Deliberation.}
Self-Refine~\citep{madaan2023selfrefine},
Reflexion~\citep{shinn2023reflexion}, and Visual
Self-Refine~\citep{li2026visualselfrefine} improve outputs through iterative
self-feedback, reflection across trials, and pixel-level visual feedback for
chart parsing, respectively. Test-time scaling offers another way to combine
different trajectories~\citep{wang2023selfconsistency,wang2025moa}.
TUMIX~\citep{chen2026tumix} combines parallel
agents with different tool-use strategies through iterative answer sharing
and refinement. Anthropic's ``think'' tool~\citep{anthropic2025think} exposes
a callable scratchpad that logs model-supplied thoughts without a new
multimodal call. \textsc{ReVision} instead executes separate tool-free
multimodal reasoning over the original image, completed interaction history,
and accumulated evidence within the same trajectory. Our registered-stub
ablation distinguishes the contribution of this interface from that of nested
execution (Section~\ref{sec:ablation}).

\section{Answer Reversals and Reasoning Compression Under Tool Access}
\label{sec:empirical}

Can tool access turn a correct tool-free answer into an incorrect one, and how does explicit reasoning change when tools are available? We investigate these questions by comparing Qwen3.5-397B-A17B with and without tools on the same 11,572 questions across six benchmarks, measuring paired answer reversals and reasoning-token use.

Let $x=(I,q)$ denote an image--question pair and let
$\mathcal{T}_{\mathrm{ext}}=\{\textsc{Search},\textsc{Visit},\textsc{Python}\}$
denote the external tools, where \textsc{Visit} opens a webpage selected from
the search results. We first compare two inference configurations
using the same underlying MLLM. In \emph{w/o Tools}, the model receives $x$
without callable external tools and produces a reasoning trace followed by a
final answer. In \emph{w/ Tools}, it receives $\mathcal{T}_{\mathrm{ext}}$ and may
interleave its own reasoning, tool calls, and observations \mbox{before answering}.

\begin{table}[t]
\centering
\small
\begin{tabular}{@{}lc@{}}
\toprule
Benchmark & \shortstack{w/o Tools Correct $\rightarrow$ w/ Tools Incorrect \\ Rate (\%)} \\
\midrule
CSVQA          & 11.18 \\
EXAMS-V        & 13.32 \\
MSEarth-MCQ    & 7.47  \\
VisualSimpleQA & 9.00  \\
WikiVQABench   & 6.40  \\
WorldVQA       & 7.27  \\
\midrule
Overall        & 9.70  \\
\bottomrule
\end{tabular}
\caption{Paired answer-reversal rates under tool access for Qwen3.5-397B-A17B. Values are the percentage of all evaluated questions answered correctly without tools but incorrectly with tools. Overall pools \mbox{11,572 questions}.}
\label{tab:answer-reversals}
\end{table}

\subsection{Aggregate Gains Hide Answer Reversals}
External tools improve Qwen3.5-397B-A17B's aggregate accuracy on all six benchmarks
(Table~\ref{tab:main}), but they do not preserve every correct tool-free answer.
Table~\ref{tab:answer-reversals} reports questions answered correctly without
tools and incorrectly with tools. These reversals motivate examining what
changes within tool-use trajectories. The qualitative comparison in
Section~\ref{sec:case-study} illustrates one way that retrieved evidence can
displace a correct \mbox{image-grounded answer}.



\begin{figure*}[!t]
    \centering
    \includegraphics[width=\textwidth]{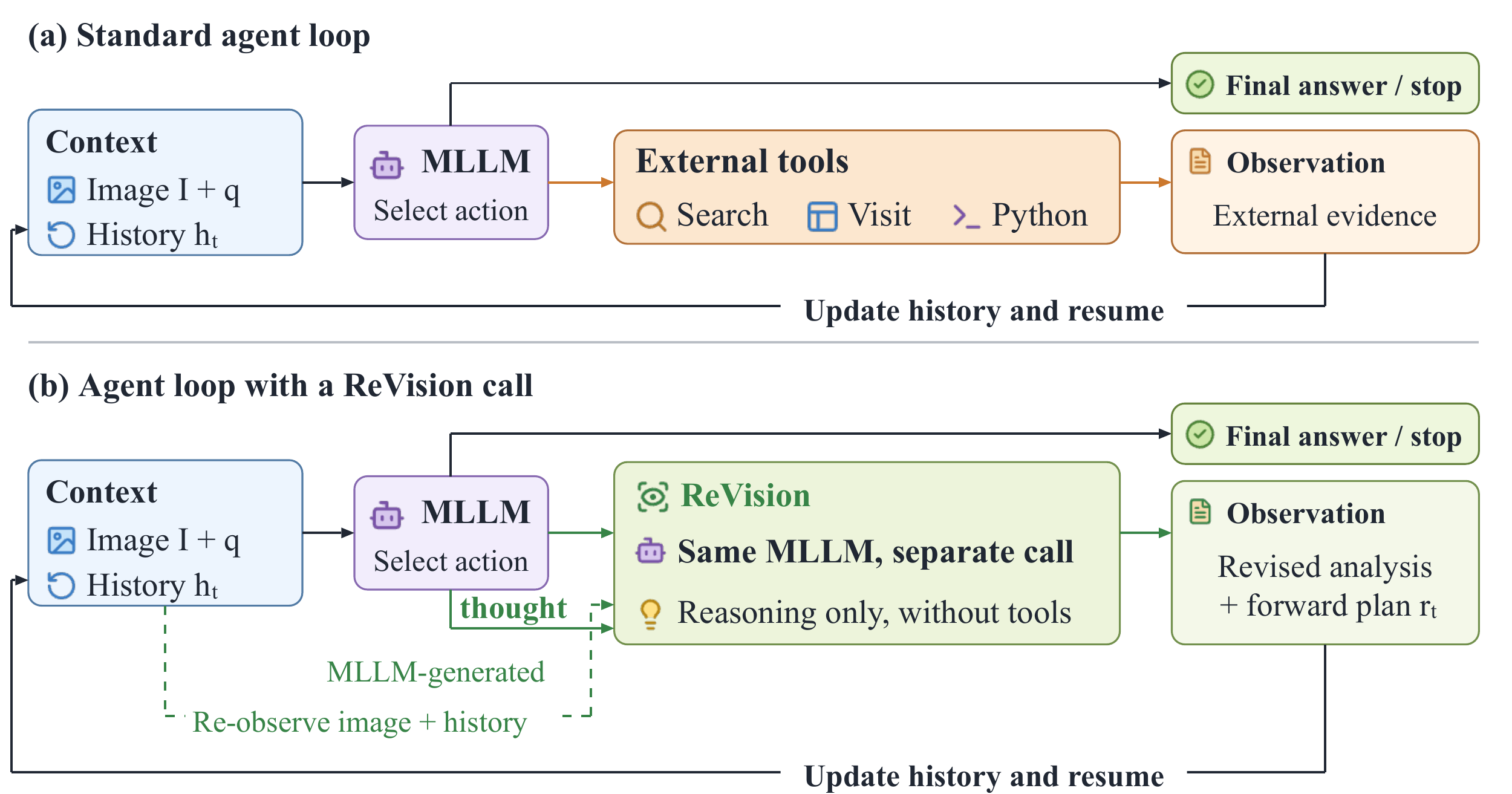}
    \caption{Method overview of standard tool use and optional \textsc{ReVision} execution. \textsc{ReVision} invokes the same backbone without external tools to revisit the original image and interaction history, then returns revised analysis and a forward plan to the MLLM. External tools remain available, and the MLLM decides when to invoke \textsc{ReVision}.}
    \label{fig:revision-method}
\end{figure*}

\subsection{Tool Access Compresses Explicit Reasoning}
Figure~\ref{fig:compression} shows that reasoning tokens per assistant turn containing reasoning
decrease on all six benchmarks by 60.5--89.8\%. Total reasoning tokens per
query also decrease on five benchmarks by 31.3--65.2\%. WikiVQABench is the
exception: its total increases by 8.6\%, despite a 72.1\% reduction per
reasoning turn. Tool access therefore changes both the length and distribution
of explicit reasoning across a trajectory. These counts measure generated
reasoning length. The case study provides a complementary view of how the
agent uses visual and retrieved evidence.

\subsection{Cross-Condition Fusion Suggests Complementary Evidence}
Dual-Rollout Combine fuses one trajectory generated with tools and one
generated without tools. It outperforms both individual inference modes on
five benchmarks (Table~\ref{tab:main}), suggesting that the two modes may
retain different useful evidence, though part of the gain could come from
additional sampling (examined in Section~\ref{sec:ablation}).
The resulting design question is whether an agent can recover useful evidence
from intrinsic reasoning during its ongoing trajectory, without always generating a second
complete rollout.

\section{ReVision: Intrinsic Reconsideration as an Optional Action}
\label{sec:method}

\begingroup\widowpenalty=10000\clubpenalty=10000
To address this question, we propose \textsc{ReVision}, which adds an optional intrinsic reconsideration action to the existing external-tool interface without additional training.\par
\endgroup

\subsection{Overview}

The resulting agent tool set is
\[
\mathcal{T}_{\mathrm{agent}}
= \mathcal{T}_{\mathrm{ext}} \cup \{\textsc{ReVision}\}.
\]
Figure~\ref{fig:revision-method} summarizes the agent's interaction loop. At each
step, the MLLM either emits a final answer or selects an action through the
standard function-calling interface. \textsc{Search}, \textsc{Visit}, and
\textsc{Python} return external evidence. \textsc{ReVision} instead executes
a separate reasoning-only call and returns its analysis as a tool observation.
The observation is appended to the history before the MLLM resumes.
\textsc{ReVision} can be skipped or invoked multiple times within a trajectory.

\subsection{Tool Interface}

\begin{figure}[!t]
    \centering
    \includegraphics[width=\columnwidth]{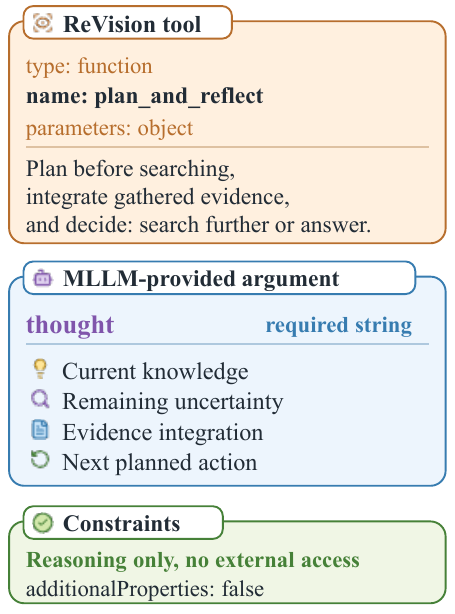}
    \caption{Function-calling interface for the optional \textsc{ReVision} action. The required \texttt{thought} argument summarizes the MLLM's current knowledge, uncertainty, evidence integration, and next planned action.}
    \label{fig:tool-schema}
\end{figure}

\textsc{ReVision} is registered as \texttt{plan\_and\_reflect} with one
required string argument, \texttt{thought} (Figure~\ref{fig:tool-schema}).
The MLLM uses this argument to record its current assessment of the
question, available evidence, uncertainty, and intended next step. The tool
description invites planning, evidence integration, and reconsideration of
the image before further search or a final answer. Invocation remains the
MLLM's decision. The runner does not schedule a reflection stage.

\subsection{Nested Reasoning Call}
Let $h_t$ denote the completed interaction history before the current action,
including prior assistant messages, tool calls, and returned observations.
When \textsc{ReVision} is selected, a separate invocation of the same
multimodal backbone receives the original image $I$, the question $q$, the
history $h_t$, and the MLLM-supplied \texttt{thought} as input. All
external tools are disabled for this call. The pending \textsc{ReVision}
action call is excluded from the nested input.
An accompanying instruction asks the model to re-examine the image, integrate the accumulated evidence, and propose the next step. The returned analysis $r_t$ is appended as the matching tool
observation. The MLLM then resumes the main trajectory and remains responsible for subsequent actions and the final answer.

\section{Experiments}

In this section, we evaluate \textsc{ReVision}'s accuracy, reasoning-token cost, the contribution of nested execution, and generalization to a second backbone, complemented by a qualitative case study.

\subsection{Setup}

\paragraph{Model and Benchmarks.}
We evaluate Qwen3.5-397B-A17B~\citep{qwen2026qwen35} on six reasoning-intensive VQA benchmarks totaling 11,572 questions.
The multiple-choice (MCQ) benchmarks are CSVQA~\citep{jian2025csvqa}
(1,378 Chinese multimodal STEM questions), EXAMS-V~\citep{das-etal-2024-exams}
(3,565 visual examination questions across languages and disciplines),
MSEarth-MCQ~\citep{zhao-etal-2026-msearth} (2,785 multimodal Earth-science
questions), and WikiVQABench~\citep{shbita2026wikivqabench} (344
knowledge-grounded visual questions using Wikipedia and Wikidata).

The open-ended QA (OpenQA) benchmarks are
VisualSimpleQA~\citep{wang2025visualsimpleqa} (500 fact-seeking questions
grounded in visual inputs) and WorldVQA~\citep{zhou2026worldvqa} (3,000
questions testing atomic real-world knowledge). Table~\ref{tab:main}
reports the corresponding evaluation metrics \mbox{and results}.

\paragraph{Baselines and Compared Method.}
We compare four configurations of the same MLLM. A \emph{rollout} is a
complete inference run. Its interaction record is a \emph{trajectory}.
\begin{itemize}
    \item \textbf{w/o Tools.} Tool-free reasoning from the image, question,
    and parametric knowledge.

    \item \textbf{w/ Tools.} Autonomous use of Search, Visit, and Python
    before answering.

    \item \textbf{Dual-Rollout Combine.} Two independently sampled
    \emph{w/o Tools} and \emph{w/ Tools} trajectories are concatenated and
    fused by an additional model call. This tests cross-condition
    complementarity at a higher fixed inference cost.

    \item \textbf{ReVision (ours).} The \emph{w/ Tools} configuration
    augmented with the intrinsic reasoning action in Section~\ref{sec:method}.
    The MLLM chooses when to invoke it, without a fixed reflection
    stage or hand-written router.
\end{itemize}

Implementation details, prompts, preprocessing, and evaluation protocols are
provided in Appendices~\ref{sec:appendix-methods}--\ref{sec:appendix-metrics}.

\begin{figure}[!t]
    \centering
    \includegraphics[width=\columnwidth]{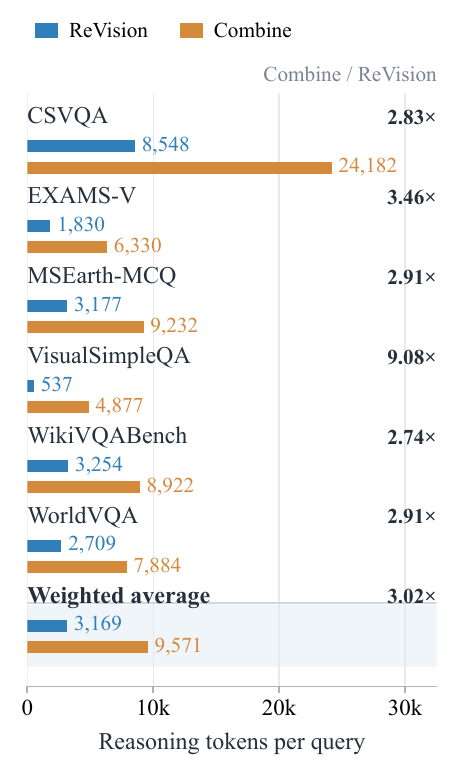}
    \caption{Reasoning-token cost comparison across inference configurations for Qwen3.5-397B-A17B. \textsc{ReVision} includes executed nested calls. Dual-Rollout Combine includes both trajectories and fusion.}
    \label{fig:combine-tokens}
\end{figure}

\begin{table*}[!t]
\centering
\begingroup
\resizebox{\textwidth}{!}{%
\begin{tabular}{llrrrrr}
\toprule
Benchmark & Metric & $N$ & w/o Tools & w/ Tools & Combine & ReVision \\
\midrule
CSVQA & MCQ Option match & 1,378 & 64.91{$\pm$0.11} & 66.57{$\pm$0.24} & 64.30{$\pm$0.73} & \textbf{67.85}{$\pm$0.94} \\
EXAMS-V & MCQ Option match & 3,565 & 65.96{$\pm$1.58} & 68.94{$\pm$3.47} & 71.30{$\pm$1.65} & \textbf{75.95}{$\pm$1.62} \\
MSEarth-MCQ & MCQ Option match & 2,785 & 65.50{$\pm$0.35} & 68.37{$\pm$0.44} & 69.39{$\pm$0.53} & \textbf{71.17}{$\pm$0.64} \\
VisualSimpleQA & Open-ended judge & 500 & 55.16{$\pm$0.96} & 65.55{$\pm$0.57} & \textbf{79.97}{$\pm$1.12} & 73.06{$\pm$0.77} \\
WikiVQABench & MCQ option match & 344 & 74.42{$\pm$0.58} & 75.98{$\pm$0.47} & 76.26{$\pm$0.36} & \textbf{77.37}{$\pm$0.61} \\
WorldVQA & Open-ended judge & 3,000 & 36.27{$\pm$1.23} & 36.32{$\pm$0.41} & 37.48{$\pm$0.89} & \textbf{40.90}{$\pm$1.58} \\
\midrule
Macro average & & 6 datasets & 60.37 & 63.62 & 66.45 & \textbf{67.72} \\
Micro average & & 11,572 & 57.81 & 60.13 & 61.76 & \textbf{64.67} \\
\bottomrule
\end{tabular}
}
\endgroup
\caption{Performance comparison on six VQA benchmarks using Qwen3.5-397B-A17B. Per-benchmark accuracy is reported as mean $\pm$ standard deviation (\%). Combine fuses one tool-free and one tool-augmented trajectory.}
\label{tab:main}
\end{table*}

\paragraph{Evaluation.}
We follow each benchmark's official evaluation protocol, including the
judge model and prompt for open-ended QA. Trajectories terminate at a final
answer or the 40-step limit. The primary Qwen3.5-397B-A17B evaluation reports
mean$\pm$std over three independent runs per condition, with matched seeds,
identical decoding settings, and the same evaluation questions. Because
answer formats and evaluators differ, we emphasize per-benchmark
improvements. The equal-weight macro average and size-weighted micro average,
$\sum_i N_i\mathrm{Acc}_i/\sum_i N_i$, provide descriptive summaries rather
than a directly comparable accuracy measure across benchmarks.

\subsection{Main Results}

\paragraph{ReVision Consistently Improves Tool Use.}

Table~\ref{tab:main} shows that tools improve accuracy on all six benchmarks, while Dual-Rollout Combine outperforms both individual modes on five. \textsc{ReVision} improves over \emph{w/o Tools} and \emph{w/ Tools} on all six, exceeds Combine on five, and achieves the highest macro and micro averages (67.72\% and 64.67\%, respectively). Its gains over \emph{w/ Tools} range from 1.28 to 7.51\%.

\subsection{Reasoning-Token Cost}

For Qwen3.5-397B-A17B, the benchmark-size-weighted mean is 3,169 reasoning tokens per
query for \textsc{ReVision}, compared with 3,023 for \emph{w/ Tools}, 4,915
for \emph{w/o Tools}, and 9,571 for Dual-Rollout Combine
(Figure~\ref{fig:combine-tokens}).

Relative to Dual-Rollout Combine, \textsc{ReVision} uses 66.9\% fewer
reasoning tokens while outperforming it on five of six benchmarks, showing
that selectively invoking intrinsic reconsideration can recover useful evidence
at substantially lower cost than generating and fusing two \mbox{complete trajectories}.


\subsection{Ablation Studies}
\label{sec:ablation}

\paragraph{Cross-Condition versus Same-Condition Fusion.}
\begin{table}[tb]
\centering
\begingroup\small
\setlength{\tabcolsep}{3pt}
\begin{tabular}{@{}lcrr@{}}
\toprule
Combination & Rollouts & \shortstack{Qwen3.5-\\397B-A17B} & Kimi-K2.5 \\
\midrule
w/ Tools & 1 & 68.37 & 65.92 \\
w/o Tools & 1 & 65.50 & 64.94 \\
w/ Tools + w/ Tools & 2 + fusion & 68.94 & 67.05 \\
w/o Tools + w/o Tools & 2 + fusion & 66.51 & 66.12 \\
w/ Tools + w/o Tools & 2 + fusion & \textbf{69.39} & \textbf{68.92} \\
\bottomrule
\end{tabular}
\endgroup
\caption{Ablation study of cross-condition versus same-condition trajectory fusion on MSEarth-MCQ. Two-rollout conditions generate two complete trajectories and use an additional fusion pass.}
\label{tab:samecondition}
\end{table}

To distinguish cross-condition complementarity from the benefit of an
additional sample, we compare \emph{w/ Tools + w/o Tools} with two
same-condition ensembles on MSEarth-MCQ. All two-rollout conditions use an
additional fusion pass. Cross-condition fusion reaches 69.39\% on Qwen3.5-397B-A17B
and 68.92\% on Kimi-K2.5, exceeding the stronger same-condition ensemble by
0.45 and 1.87\%, respectively (Table~\ref{tab:samecondition}).
This comparison supports a contribution from mixing inference modes on this
benchmark, alongside the benefit of multiple trajectories.

\paragraph{Alternative Reasoning Strategies.}
\begin{table}[tb]
\centering
\begingroup
\setlength{\tabcolsep}{2.8pt}
\resizebox{\columnwidth}{!}{%
\begin{tabular}{@{}lrrr@{}}
\toprule
Method & \shortstack{MSEarth-\\MCQ} & \shortstack{Visual\\SimpleQA} & WorldVQA \\
\midrule
w/ Tools & 68.37 & 65.55 & 36.32 \\
Think scratchpad & 69.34 & 61.53 & 26.61 \\
Prompt think more & 68.60 & 64.43 & 30.73 \\
Force think / 3 & 68.54 & 62.49 & 34.09 \\
\textsc{ReVision} & \textbf{71.17} & \textbf{73.06} & \textbf{40.90} \\
\bottomrule
\end{tabular}
}
\endgroup
\caption{Ablation study of alternative reasoning strategies on three VQA benchmarks. Values are \mbox{accuracy (\%)}.}
\label{tab:reasoning-ablation}
\end{table}

Table~\ref{tab:reasoning-ablation} compares \textsc{ReVision} with three alternative ways to elicit additional reasoning: (1) Think scratchpad: an optional action modeled on Anthropic's ``Think'' tool~\citep{anthropic2025think} that returns the model-supplied thought without executing a separate reasoning call, (2) Prompt think more: a prompt-only instruction to reason more before acting, and (3) Periodic forced thinking: a reasoning step enforced after every three agent steps rather than selected by the model. All three alternatives improve over \emph{w/ Tools} on MSEarth-MCQ but underperform it on VisualSimpleQA and WorldVQA. \textsc{ReVision} has the highest accuracy on all three benchmarks.

\paragraph{Interface Registration versus Nested Execution.}
\begin{table}[!tb]
\centering
\begingroup
\setlength{\tabcolsep}{2.4pt}
\resizebox{\columnwidth}{!}{%
\begin{tabular}{@{}lrrr@{}}
\toprule
Method & \shortstack{Visual\\SimpleQA} & \shortstack{WikiVQA\\Bench} & WorldVQA \\
\midrule
w/ Tools & 65.55 & 75.98 & 36.32 \\
Registered stub & 71.00 & 75.29 & 37.73 \\
\textsc{ReVision} & \textbf{73.06} & \textbf{77.37} & \textbf{40.90} \\
\bottomrule
\end{tabular}
}
\endgroup
\caption{Ablation study of \textsc{ReVision} interface registration and nested reasoning execution. The registered stub returns a placeholder instead of executing the nested call. Values are accuracy (\%).}
\label{tab:noop-ablation}
\end{table}

Table~\ref{tab:noop-ablation} tests the contributions of interface registration
and nested execution. The registered stub keeps the MLLM system prompt
and \textsc{ReVision} interface fixed but returns a placeholder instead of
running the call. On VisualSimpleQA and WorldVQA the stub already improves over \emph{w/ Tools}
(by 5.45 and 1.41 points, respectively), suggesting that making the reasoning
action available can itself steer the MLLM. On WikiVQABench it slightly
underperforms ($-$0.69). \textsc{ReVision} outperforms the stub on all three
benchmarks, showing that executing the image-grounded reasoning step adds a
consistent benefit beyond interface \mbox{registration alone}.

\subsection{Generalization to Different Models}

\begin{table}[tb]
\centering
\begingroup\small
\setlength{\tabcolsep}{2.6pt}
\begin{tabular}{@{}lrrr@{}}
\toprule
Method / measure & EXAMS-V & \shortstack{MSEarth-\\MCQ} & \shortstack{Visual\\SimpleQA} \\
\midrule
w/o Tools (Acc.) & 75.85 & 64.94 & 57.39 \\
w/ Tools (Acc.) & 75.71 & 65.92 & 66.71 \\
\textsc{ReVision} (Acc.) & \textbf{78.12} & \textbf{68.22} & \textbf{69.32} \\
\midrule
w/ Tools (Tokens) & 2,402 & 2,762 & 651 \\
\textsc{ReVision} (Tokens) & 1,917 & 2,029 & 428 \\
w/o Tools (Tokens) & 3,170 & 3,311 & 1,486 \\
\bottomrule
\end{tabular}
\endgroup
\caption{Kimi-K2.5 results on three benchmarks: accuracy (\%) and reasoning tokens per query.}
\label{tab:kimi-transfer}
\end{table}

We repeat the three single-trajectory conditions on Kimi-K2.5~\citep{kimi2026k25} over EXAMS-V,
MSEarth-MCQ, and VisualSimpleQA, using one run per condition. The same
\textsc{ReVision} interface invokes a multimodal reasoning call without
tools. \textsc{ReVision} improves over \emph{w/ Tools} by 2.41, 2.30, and
2.61\%, respectively, while reducing reasoning tokens on all
three benchmarks (Table~\ref{tab:kimi-transfer}). These results extend the
observed benefit to a second backbone. The primary three-run evidence remains
the Qwen3.5-397B-A17B evaluation.

\begin{figure}[!b]
    \centering
    \includegraphics[width=\columnwidth]{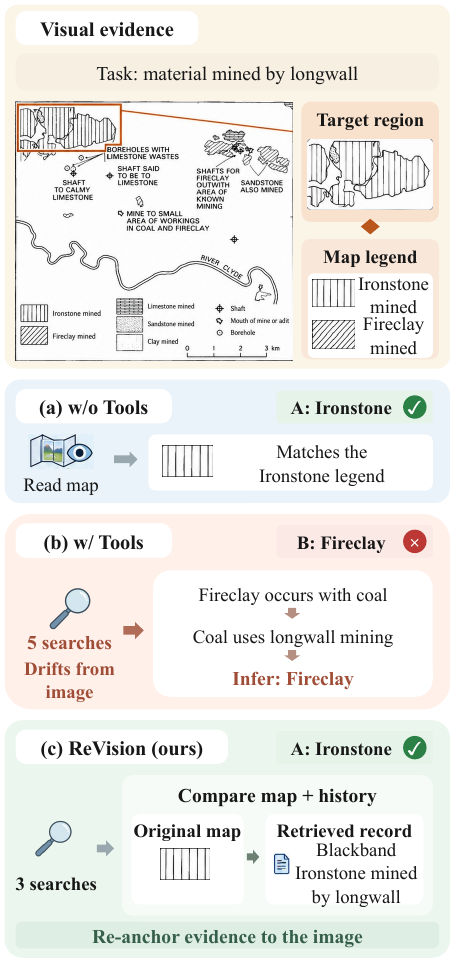}
    \caption{Qualitative comparison of tool-free reasoning, standard tool use, and \textsc{ReVision} on an MSEarth-MCQ question. \textsc{ReVision} checks retrieved evidence against the map and restores the correct answer.}
    \label{fig:revision-case}
\end{figure}

\subsection{Qualitative Case Study}
\label{sec:case-study}

Figure~\ref{fig:revision-case} compares the three inference conditions on the same MSEarth-MCQ question. On this
MSEarth-MCQ item, \emph{w/o Tools} maps the large continuous region labeled
``Ironstone mined'' to longwall extraction and answers A. Standard \emph{w/ Tools}
instead makes five searches and follows an indirect shortcut---fireclay
occurs with coal, and coal is commonly mined by longwall---to answer B.
In the \textsc{ReVision} trajectory, the MLLM autonomously invokes reconsideration after three searches.
The nested call re-observes the map and compares it with a
retrieved record that Blackband Ironstone was worked by longwall, restoring
the correct answer A. Thus, the correction keeps the retrieved evidence but
re-anchors it to the image.

\section{Discussion}
\label{sec:discussion}

\subsection{Preserving Evidence within a Trajectory}
The observations reveal a tension between aggregate tool benefits and the
preservation of correct tool-free solutions. Reasoning turns become shorter
under tool access, while some previously correct answers are lost. The case
study illustrates one possible failure: external associations redirect the
answer away from the relevant visual evidence. \textsc{ReVision} provides a
point at which the agent can reconsider that evidence together with its
accumulated observations. Reasoning length and answer reversals characterize
the observed behavioral shift under tool access.

\subsection{Where Additional Reasoning Helps}
The ablations separate two contributions: registering a reasoning action can
steer the MLLM, and executing the nested call adds further gains.
Meanwhile, cross-condition fusion can recover information from independently
generated trajectories, but requires two complete rollouts and a fusion pass.
\textsc{ReVision} offers a different allocation of inference effort:
reconsideration is available within the ongoing trajectory and invoked by the
MLLM. Its advantage is task-dependent, as the stronger VisualSimpleQA
result from Combine makes clear.
The invocation decision can use evidence unavailable at the start of the trajectory, letting the agent revisit an earlier interpretation as new observations arrive.

\section{Conclusion}

In this paper, we study how tool-augmented MLLMs coordinate intrinsic visual reasoning with external evidence in reasoning-intensive VQA. Through our empirical study across six benchmarks, we find that tool access improves aggregate accuracy but can also reverse correct tool-free answers and substantially shorten explicit reasoning turns. Based on these findings, we propose \textsc{ReVision}, a training-free test-time method that makes intrinsic reconsideration an optional action alongside external tools. The MLLM decides when to invoke a separate tool-free multimodal call to re-examine the original image and integrate accumulated evidence. Experiments show that \textsc{ReVision} improves accuracy over standard tool use by 1.28--7.51\% across all six benchmarks, with comparable reasoning-token use. It also outperforms Dual-Rollout Combine on five benchmarks while using substantially fewer reasoning tokens. These results highlight the value of selectively revisiting intrinsic reasoning during tool use and suggest the potential of \textsc{ReVision} as a general approach to coordinating intrinsic reasoning and external evidence in broader agent settings.

\section*{Limitations}

Our evaluation covers reasoning-intensive multimodal VQA across six
benchmarks and two backbones. The core mechanism---re-examining a primary
input to reconcile externally acquired evidence---is not specific to images
or VQA, but we have not yet tested \textsc{ReVision} on text-only agent
tasks or long-horizon interactive environments. Because the method is
training-free and requires only registering an additional action, applying it
to new settings involves no architectural change. We leave such extensions to
future work.

Our analysis uses reasoning-token counts to characterize a behavioral shift
under tool access, not to establish a causal link between shorter reasoning
and incorrect answers. The effectiveness of \textsc{ReVision} is
demonstrated through direct accuracy comparisons and controlled ablations,
independently of the causal status of reasoning compression. In particular,
the ablation separating interface registration from nested execution
(Table~\ref{tab:noop-ablation}) and the cross-condition versus
same-condition fusion comparison (Table~\ref{tab:samecondition}) provide
evidence that goes beyond aggregate token statistics.

\textsc{ReVision} relies on the MLLM to decide when to invoke
reconsideration, without a learned invocation policy. This training-free design is a deliberate choice: it avoids task-specific supervision and supports reuse across backbones, as illustrated by the Kimi-K2.5 results. Learning when to invoke reconsideration is a complementary direction that could further improve the current training-free method.

\section*{Potential Risks}

\textsc{ReVision} is intended to support accurate visual question
answering. Its ability to revisit visual details using accumulated
external evidence could also be applied to intrusive tasks, such as
inferring where a person lives or which places they frequent from
photographs without their consent. Agents incorporating the method could
also be repurposed to generate misleading descriptions of visual content
for disinformation. Responsible deployment should respect permissions
for image and data use and restrict applications that violate privacy
or intentionally deceive others.

\bibliography{references}

@misc{jian2025csvqa,
  archiveprefix = {arXiv},
  author = {Ai Jian and Weijie Qiu and Xiaokun Wang and Peiyu Wang and Yunzhuo Hao and Jiangbo Pei and Yichen Wei and Yi Peng and Xuchen Song},
  doi = {10.48550/arXiv.2505.24120},
  eprint = {2505.24120},
  primaryclass = {cs.CV},
  title = {{CSVQA}: A {Chinese} Multimodal Benchmark for Evaluating {STEM} Reasoning Capabilities of {VLMs}},
  url = {https://arxiv.org/abs/2505.24120},
  year = {2025}
}

@inproceedings{das-etal-2024-exams,
  address = {Bangkok, Thailand},
  author = {Das, Rocktim  and
Hristov, Simeon  and
Li, Haonan  and
Dimitrov, Dimitar  and
Koychev, Ivan  and
Nakov, Preslav},
  booktitle = {Proceedings of the 62nd Annual Meeting of the {A}ssociation for {C}omputational {L}inguistics (Volume 1: Long Papers)},
  doi = {10.18653/v1/2024.acl-long.420},
  editor = {Ku, Lun-Wei  and
Martins, Andre  and
Srikumar, Vivek},
  month = {August},
  pages = {7768--7791},
  publisher = {Association for Computational Linguistics},
  title = {{EXAMS}-{V}: A Multi-Discipline Multilingual Multimodal Exam Benchmark for Evaluating Vision Language Models},
  url = {https://aclanthology.org/2024.acl-long.420/},
  year = {2024}
}

@inproceedings{zhao-etal-2026-msearth,
  address = {San Diego, California, United States},
  author = {Zhao, Xiangyu  and
Xu, Wanghan  and
Liu, Bo  and
Zhou, Yuhao  and
Ling, Fenghua  and
Fei, Ben  and
Yue, Xiaoyu  and
Bai, Lei  and
Zhang, Wenlong  and
Wu, Xiao-Ming},
  booktitle = {Proceedings of the 64th Annual Meeting of the {A}ssociation for {C}omputational {L}inguistics (Volume 1: Long Papers)},
  doi = {10.18653/v1/2026.acl-long.239},
  editor = {Liakata, Maria  and
Moreira, Viviane P.  and
Zhang, Jiajun  and
Jurgens, David},
  isbn = {979-8-89176-390-6},
  month = {July},
  pages = {5270--5301},
  publisher = {Association for Computational Linguistics},
  title = {{MSE}arth: A Multimodal Benchmark for {Earth} Science Phenomenon Discovery with {MLLM}s},
  url = {https://aclanthology.org/2026.acl-long.239/},
  year = {2026}
}

@misc{wang2025visualsimpleqa,
  archiveprefix = {arXiv},
  author = {Yanling Wang and Yihan Zhao and Xiaodong Chen and Shasha Guo and Lixin Liu and Haoyang Li and Yong Xiao and Jing Zhang and Qi Li and Ke Xu},
  doi = {10.48550/arXiv.2503.06492},
  eprint = {2503.06492},
  primaryclass = {cs.CL},
  title = {{VisualSimpleQA}: A Benchmark for Decoupled Evaluation of Large Vision-Language Models in Fact-Seeking Question Answering},
  url = {https://arxiv.org/abs/2503.06492},
  year = {2025}
}

@misc{shbita2026wikivqabench,
  archiveprefix = {arXiv},
  author = {Basel Shbita and Pengyuan Li and Anna Lisa Gentile},
  doi = {10.48550/arXiv.2605.21479},
  eprint = {2605.21479},
  primaryclass = {cs.CV},
  title = {{WikiVQABench}: A Knowledge-Grounded Visual Question Answering Benchmark from {Wikipedia} and {Wikidata}},
  url = {https://arxiv.org/abs/2605.21479},
  year = {2026}
}

@misc{zhou2026worldvqa,
  archiveprefix = {arXiv},
  author = {Runjie Zhou and Youbo Shao and Haoyu Lu and Bowei Xing and Tongtong Bai and Yujie Chen and Jie Zhao and Lin Sui and Haotian Yao and Zijia Zhao and Hao Yang and Haoning Wu and Zaida Zhou and Jinguo Zhu and Zhiqi Huang and Yiping Bao and Yangyang Liu and Y. Charles and Xinyu Zhou},
  doi = {10.48550/arXiv.2602.02537},
  eprint = {2602.02537},
  primaryclass = {cs.CV},
  title = {{WorldVQA}: Measuring Atomic World Knowledge in Multimodal Large Language Models},
  url = {https://arxiv.org/abs/2602.02537},
  year = {2026}
}

@inproceedings{yao2023react,
  author = {Yao, Shunyu and Zhao, Jeffrey and Yu, Dian and Du, Nan and Shafran, Izhak and Narasimhan, Karthik and Cao, Yuan},
  booktitle = {International Conference on Learning Representations (ICLR) },
  html = {https://arxiv.org/abs/2210.03629},
  title = {{ReAct}: Synergizing Reasoning and Acting in Language Models},
  year = {2023}
}

@misc{yang2023mmreact,
  archiveprefix = {arXiv},
  author = {Zhengyuan Yang and Linjie Li and Jianfeng Wang and Kevin Lin and Ehsan Azarnasab and Faisal Ahmed and Zicheng Liu and Ce Liu and Michael Zeng and Lijuan Wang},
  doi = {10.48550/arXiv.2303.11381},
  eprint = {2303.11381},
  primaryclass = {cs.CV},
  title = {{MM-REACT}: Prompting {ChatGPT} for Multimodal Reasoning and Action},
  url = {https://arxiv.org/abs/2303.11381},
  year = {2023}
}

@inproceedings{lu2023chameleon,
  author = {Lu, Pan and Peng, Baolin and Cheng, Hao and Galley, Michel and Chang, Kai-Wei and Wu, Ying Nian and Zhu, Song-Chun and Gao, Jianfeng},
  booktitle = {Advances in Neural Information Processing Systems 36},
  collection = {NeurIPS 2023},
  doi = {10.52202/075280-1882},
  pages = {43447--43478},
  publisher = {Neural Information Processing Systems Foundation, Inc. (NeurIPS)},
  series = {NeurIPS 2023},
  title = {{Chameleon}: Plug-and-Play Compositional Reasoning with Large Language Models},
  url = {http://dx.doi.org/10.52202/075280-1882},
  year = {2023}
}

@inproceedings{liu2025llavaplus,
  address = {Cham},
  author = {Liu, Shilong
and Cheng, Hao
and Liu, Haotian
and Zhang, Hao
and Li, Feng
and Ren, Tianhe
and Zou, Xueyan
and Yang, Jianwei
and Su, Hang
and Zhu, Jun
and Zhang, Lei
and Gao, Jianfeng
and Li, Chunyuan},
  booktitle = {Computer Vision -- ECCV 2024},
  doi = {10.1007/978-3-031-72970-6_8},
  editor = {Leonardis, Ale{\v{s}}
and Ricci, Elisa
and Roth, Stefan
and Russakovsky, Olga
and Sattler, Torsten
and Varol, G{\"u}l},
  isbn = {978-3-031-72970-6},
  pages = {126--142},
  publisher = {Springer Nature Switzerland},
  title = {{LLaVA-Plus}: Learning to Use Tools for Creating Multimodal Agents},
  year = {2025}
}

@inproceedings{qian2025smart,
  address = {Vienna, Austria},
  author = {Qian, Cheng  and
Acikgoz, Emre Can  and
Wang, Hongru  and
Chen, Xiusi  and
Sil, Avirup  and
Hakkani-T{\"u}r, Dilek  and
Tur, Gokhan  and
Ji, Heng},
  booktitle = {Findings of the {A}ssociation for {C}omputational {L}inguistics: {ACL} 2025},
  doi = {10.18653/v1/2025.findings-acl.239},
  editor = {Che, Wanxiang  and
Nabende, Joyce  and
Shutova, Ekaterina  and
Pilehvar, Mohammad Taher},
  isbn = {979-8-89176-256-5},
  month = {July},
  pages = {4604--4621},
  publisher = {Association for Computational Linguistics},
  title = {{SMART}: Self-Aware Agent for Tool Overuse Mitigation},
  url = {https://aclanthology.org/2025.findings-acl.239/},
  year = {2025}
}

@inproceedings{li2025meco,
  address = {Vienna, Austria},
  author = {Li, Wenjun  and
Li, Dexun  and
Dong, Kuicai  and
Zhang, Cong  and
Zhang, Hao  and
Liu, Weiwen  and
Wang, Yasheng  and
Tang, Ruiming  and
Liu, Yong},
  booktitle = {Proceedings of the 63rd Annual Meeting of the {A}ssociation for {C}omputational {L}inguistics (Volume 1: Long Papers)},
  doi = {10.18653/v1/2025.acl-long.655},
  editor = {Che, Wanxiang  and
Nabende, Joyce  and
Shutova, Ekaterina  and
Pilehvar, Mohammad Taher},
  isbn = {979-8-89176-251-0},
  month = {July},
  pages = {13346--13370},
  publisher = {Association for Computational Linguistics},
  title = {Adaptive Tool Use in Large Language Models with Meta-Cognition Trigger},
  url = {https://aclanthology.org/2025.acl-long.655/},
  year = {2025}
}

@inproceedings{wang2025toa,
  author = {Wang, Hongru and Qian, Cheng and Li, Manling and Qiu, Jiahao and Xue, Boyang and Wang, Mengdi and Ji, Heng and Storkey, Amos and Wong, Kam-Fai},
  booktitle = {Proceedings of the 43rd International Conference on Machine Learning},
  editor = {Zhang, Tong and Dudik, Miroslav and Jaggi, Martin and Agarwal, Alekh and Li, Sharon and Schuurmans, Dale and Zhu, Jerry and Berkenkamp, Felix and Dong, Hanze and Bietti, Alberto},
  month = {06--11 Jul},
  pages = {172470--172485},
  publisher = {PMLR},
  series = {Proceedings of Machine Learning Research},
  title = {Position: Agents Should Invoke External Tools {ONLY} When Epistemically Necessary},
  url = {https://proceedings.mlr.press/v306/wang26mb.html},
  volume = {306},
  year = {2026}
}

@inproceedings{deng2026toolscope,
  address = {San Diego, California, United States},
  author = {Deng, Mengjie  and
Dong, Guanting  and
Dou, Zhicheng},
  booktitle = {Findings of the {A}ssociation for {C}omputational {L}inguistics: {ACL} 2026},
  doi = {10.18653/v1/2026.findings-acl.11},
  editor = {Liakata, Maria  and
Moreira, Viviane P.  and
Zhang, Jiajun  and
Jurgens, David},
  isbn = {979-8-89176-395-1},
  month = {July},
  pages = {211--225},
  publisher = {Association for Computational Linguistics},
  title = {{T}ool{S}cope: An Agentic Framework for Vision-Guided and Long-Horizon Tool Use},
  url = {https://aclanthology.org/2026.findings-acl.11/},
  year = {2026}
}

@misc{guo2026multimodaltools,
  archiveprefix = {arXiv},
  author = {Jiawei Guo and Donglei Yu and Yu Chen and Xiang Wang and Shuai Li and Xinpei Zhao and Huaxing Liu and Qinghao Wang and Minpeng Liao},
  doi = {10.48550/arXiv.2606.02357},
  eprint = {2606.02357},
  primaryclass = {cs.CV},
  title = {Do Multimodal Agents Really Benefit from Tool Use? {A} Systematic Study of Capability Gains},
  url = {https://arxiv.org/abs/2606.02357},
  year = {2026}
}

@inproceedings{chen2026tumix,
  author = {Chen, Yongchao and Chen, Jiefeng and Meng, Rui and Yin, Ji and Li, Na and Fan, Chuchu and Wang, Chi and Pfister, Tomas and Yoon, Jinsung},
  booktitle = {International Conference on Learning Representations},
  editor = {C. Vondrick and B. Hariharan and C. Raffel and L. Pinto and D. Yang and A. Faust},
  pages = {55848--55878},
  title = {{TUMIX}: Multi-Agent Test-Time Scaling with Tool-Use Mixture},
  url = {https://proceedings.iclr.cc/paper_files/paper/2026/file/5b755cf5598a4324d253025e1fbbba52-Paper-Conference.pdf},
  volume = {2026},
  year = {2026}
}

@misc{zeng2026tooloveruse,
  archiveprefix = {arXiv},
  author = {Yirong Zeng and Shen You and Yufei Liu and Qunyao Du and Xiao Ding and Yutai Hou and Yuxian Wang and Wu Ning and Haonan Song and Dandan Tu and Bibo Cai and Ting Liu},
  doi = {10.48550/arXiv.2604.19749},
  eprint = {2604.19749},
  primaryclass = {cs.AI},
  title = {The Tool-Overuse Illusion: Why Does {LLM} Prefer External Tools over Internal Knowledge?},
  url = {https://arxiv.org/abs/2604.19749},
  year = {2026}
}

@inproceedings{madaan2023selfrefine,
  author = {Madaan, Aman and Tandon, Niket and Gupta, Prakhar and Hallinan, Skyler and Gao, Luyu and Wiegreffe, Sarah and Alon, Uri and Dziri, Nouha and Prabhumoye, Shrimai and Yang, Yiming and Gupta, Shashank and Majumder, Bodhisattwa Prasad and Hermann, Katherine and Welleck, Sean and Yazdanbakhsh, Amir and Clark, Peter},
  booktitle = {Advances in Neural Information Processing Systems 36},
  collection = {NeurIPS 2023},
  doi = {10.52202/075280-2019},
  pages = {46534--46594},
  publisher = {Neural Information Processing Systems Foundation, Inc. (NeurIPS)},
  series = {NeurIPS 2023},
  title = {{Self-Refine}: Iterative Refinement with Self-Feedback},
  url = {http://dx.doi.org/10.52202/075280-2019},
  year = {2023}
}

@inproceedings{shinn2023reflexion,
  author = {Shinn, Noah and Cassano, Federico and Gopinath, Ashwin and Narasimhan, Karthik and Yao, Shunyu},
  booktitle = {Advances in Neural Information Processing Systems 36},
  collection = {NeurIPS 2023},
  doi = {10.52202/075280-0377},
  pages = {8634--8652},
  publisher = {Neural Information Processing Systems Foundation, Inc. (NeurIPS)},
  series = {NeurIPS 2023},
  title = {{Reflexion}: language agents with verbal reinforcement learning},
  url = {http://dx.doi.org/10.52202/075280-0377},
  year = {2023}
}

@misc{anthropic2025think,
  author = {{Anthropic}},
  howpublished = {Anthropic Engineering},
  month = {mar},
  note = {Accessed: 2026-10-13},
  title = {The ``Think'' Tool: Enabling {Claude} to Stop and Think in Complex Tool Use Situations},
  url = {https://www.anthropic.com/engineering/claude-think-tool},
  year = {2025}
}

@inproceedings{zheng2026deepeyes,
  author = {Zheng, Ziwei and Yang, Minghao and Hong, Jack and Zhao, Chenxiao and Xu, Guohai and Yang, Le and Shen, Chao and XingYu},
  booktitle = {International Conference on Learning Representations},
  editor = {C. Vondrick and B. Hariharan and C. Raffel and L. Pinto and D. Yang and A. Faust},
  pages = {126775--126798},
  title = {{DeepEyes}: Incentivizing "Thinking with Images" via Reinforcement Learning},
  url = {https://proceedings.iclr.cc/paper_files/paper/2026/file/cdb347d7516d52a9280ea9d6708911f6-Paper-Conference.pdf},
  volume = {2026},
  year = {2026}
}

@inproceedings{wu2026mmsearchr1,
  address = {San Diego, California, United States},
  author = {Wu, Jinming  and
Deng, Zihao  and
Li, Wei  and
Liu, Yiding  and
You, Bo  and
Li, Bo  and
MA, Zejun  and
Liu, Ziwei},
  booktitle = {Proceedings of the 64th Annual Meeting of the {A}ssociation for {C}omputational {L}inguistics (Volume 1: Long Papers)},
  doi = {10.18653/v1/2026.acl-long.114},
  editor = {Liakata, Maria  and
Moreira, Viviane P.  and
Zhang, Jiajun  and
Jurgens, David},
  isbn = {979-8-89176-390-6},
  month = {July},
  pages = {2456--2487},
  publisher = {Association for Computational Linguistics},
  title = {{MMS}earch-{R}1: Incentivizing {LMM}s to Search},
  url = {https://aclanthology.org/2026.acl-long.114/},
  year = {2026}
}

@inproceedings{lou2026toolillusion,
  author = {Lou, Renze and Peng, Baolin and Yao, Wenlin and Wu, Qianhui and Cheng, Hao and Nath, Suman and Yin, Wenpeng and Gao, Jianfeng},
  booktitle = {2026 COLM},
  month = {July},
  title = {The Tool Illusion: Rethinking Tool Use in Web Agents},
  url = {https://www.microsoft.com/en-us/research/publication/the-tool-illusion-rethinking-tool-use-in-web-agents/},
  year = {2026}
}

@inproceedings{ma2026autotool,
  author = {Ma, Qinghe and Zhao, Zhen and Wu, Yiming and Zhang, Jian and Bai, Lei and Shi, Yinghuan},
  booktitle = {Proceedings of the 43rd International Conference on Machine Learning},
  editor = {Zhang, Tong and Dudik, Miroslav and Jaggi, Martin and Agarwal, Alekh and Li, Sharon and Schuurmans, Dale and Zhu, Jerry and Berkenkamp, Felix and Dong, Hanze and Bietti, Alberto},
  month = {06--11 Jul},
  pages = {85364--85389},
  publisher = {PMLR},
  series = {Proceedings of Machine Learning Research},
  title = {Are Tools Always Beneficial? {L}earning to Invoke Tools Adaptively for Dual-Mode Multimodal {LLM} Reasoning},
  url = {https://proceedings.mlr.press/v306/ma26ay.html},
  volume = {306},
  year = {2026}
}

@inproceedings{li2026visualselfrefine,
  author = {Li, Jinsong and Dong, Xiaoyi and Zang, Yuhang and Cao, Yuhang and Wang, Jiaqi and Lin, Dahua},
  booktitle = {International Conference on Learning Representations},
  editor = {C. Vondrick and B. Hariharan and C. Raffel and L. Pinto and D. Yang and A. Faust},
  pages = {85499--85522},
  title = {Visual {Self-Refine}: A Pixel-Guided Paradigm for Accurate Chart Parsing},
  url = {https://proceedings.iclr.cc/paper_files/paper/2026/file/89b0e466b46292ce0bfe53618aadd3de-Paper-Conference.pdf},
  volume = {2026},
  year = {2026}
}

@misc{qwen2026qwen35,
  author = {{Qwen Team}},
  month = {February},
  title = {{Qwen3.5}: Towards Native Multimodal Agents},
  url = {https://qwen.ai/blog?id=qwen3.5},
  year = {2026}
}

@misc{kimi2026k25,
  archiveprefix = {arXiv},
  author = {{Kimi Team} and Tongtong Bai and Yifan Bai and Yiping Bao and S. H. Cai and Yuan Cao and Ziwei Chai and Y. Charles and H. S. Che and Cheng Chen and Guanduo Chen and Huarong Chen and Jia Chen and Jianlong Chen and Jun Chen and Kefan Chen and Liang Chen and Ruijue Chen and Xinhao Chen and Yanru Chen and Yanxu Chen and Yicun Chen and Yimin Chen and Yingjiang Chen and Yuankun Chen and Yujie Chen and Yutian Chen and Zhirong Chen and Ziwei Chen and Dazhi Cheng and Yean Cheng and Minghan Chu and Jialei Cui and Jiaqi Deng and Muxi Diao and Hao Ding and Mengfan Dong and Mengnan Dong and Yuxin Dong and Yuhao Dong and Angang Du and Chenzhuang Du and Dikang Du and Lingxiao Du and Yulun Du and Yu Fan and Shengjun Fang and Qiulin Feng and Yichen Feng and Garimugai Fu and Kelin Fu and Hongcheng Gao and Tong Gao and Yuyao Ge and Shangyi Geng and Chengyang Gong and Xiaochen Gong and Zhuoma Gongque and Qizheng Gu and Xinran Gu and Yicheng Gu and Longyu Guan and Shuhao Guan and Yuanying Guo and Xiaoru Hao and Dailan He and Tianhong He and Weiran He and Wenyang He and Yibo He and Yunjia He and Chao Hong and Hao Hu and Jiaxi Hu and Yangyang Hu and Zhenxing Hu and Ke Huang and Ruiyuan Huang and Weixiao Huang and Zhiqi Huang and Chaobo Jia and Tao Jiang and Zhejun Jiang and Xinyi Jin and Yu Jing and Guokun Lai and Aidi Li and C. Li and Cheng Li and Fang Li and Guanghe Li and Guanyu Li and Haitao Li and Haoyang Li and Jia Li and Jingwei Li and Junxiong Li and Lincan Li and Mo Li and Weihong Li and Wentao Li and Xinhang Li and Xinhao Li and Yang Li and Yanhao Li and Yiwei Li and Yuxiao Li and Zhaowei Li and Zhaoxi Li and Zheming Li and Weilong Liao and Jiawei Lin and Xiaohan Lin and Yibo Lin and Zhishan Lin and Zichao Lin and Cheng Liu and Chenyu Liu and Hongzhang Liu and Liang Liu and Shaowei Liu and Shudong Liu and Shuran Liu and Tianwei Liu and Tianyu Liu and Weizhou Liu and Xiangyan Liu and Yangyang Liu and Yanming Liu and Yibo Liu and Yuanxin Liu and Zhengying Liu and Zhongnuo Liu and Enzhe Lu and Haoyu Lu and Zhiyuan Lu and G. Luo and Junyu Luo and Tongxu Luo and Yashuo Luo and Long Ma and Shaoguang Mao and Yuan Mei and Xin Men and Fanqing Meng and Zhiyong Meng and Yibo Miao and Minqing Ni and Kun Ouyang and Siyuan Pan and Bo Pang and Yuchao Qian and Ruoyu Qin and Zeyu Qin and Jiezhong Qiu and Bowen Qu and Zeyu Shang and Youbo Shao and Tianxiao Shen and Zhennan Shen and Juanfeng Shi and Lidong Shi and Shengyuan Shi and Feifan Song and Pengwei Song and Tianhui Song and Xiaoxi Song and Hongjin Su and Jianlin Su and Zhaochen Su and Lin Sui and Jinsong Sun and Junyao Sun and Tongyu Sun and Flood Sung and Yunpeng Tai and Chuning Tang and Heyi Tang and Xiaojuan Tang and Zhengyang Tang and Jiawen Tao and Shiyuan Teng and Chaoran Tian and Pengfei Tian and Bowen Wang and Chensi Wang and Chuang Wang and Congcong Wang and Dingkun Wang and Dinglu Wang and Dongliang Wang and Feng Wang and Hailong Wang and Haiming Wang and Hao Wang and Hengzhi Wang and Huaqing Wang and Hui Wang and Jiahao Wang and Jinhong Wang and Jiuzheng Wang and Kaixin Wang and Linian Wang and Qibin Wang and Shengjie Wang and Shuyi Wang and Si Wang and Wei Wang and Xiaochen Wang and Xinyuan Wang and Yao Wang and Yejie Wang and Yipu Wang and Yiqin Wang and Yucheng Wang and Yuzhi Wang and Zhaoji Wang and Zhaowei Wang and Zhengtao Wang and Zhexu Wang and Zifan Wang and Zihan Wang and Zizhe Wang and Chu Wei and Ming Wei and Chuan Wen and Zichen Wen and Chengjie Wu and Haoning Wu and Junyan Wu and Rucong Wu and Wenhao Wu and Yuefeng Wu and Yuhao Wu and Yuxin Wu and Zijian Wu and Chenjun Xiao and Jin Xie and Xiaotong Xie and Yuchong Xie and Bowei Xing and Boyu Xu and Jianfan Xu and Jing Xu and Jinjing Xu and L. H. Xu and Lin Xu and Suting Xu and Weixin Xu and Xinbo Xu and Xinran Xu and Yangchuan Xu and Yichang Xu and Yuemeng Xu and Zelai Xu and Ziyao Xu and Junjie Yan and Yuzi Yan and Guangyao Yang and Hao Yang and Junwei Yang and Kai Yang and Ningyuan Yang and Xiaofei Yang and Xinlong Yang and Xinyu Yang and Ying Yang and Yi Yang and Yi Yang and Zhen Yang and Zhilin Yang and Zonghan Yang and Haotian Yao and Dan Ye and Haoran Ye and Wenjie Ye and Zhuorui Ye and Peng Yebo and Bohong Yin and Chengzhen Yu and Longhui Yu and Tao Yu and Tianxiang Yu and Enming Yuan and Mengjie Yuan and Xiaokun Yuan and Yang Yue and Weihao Zeng and Dunyuan Zha and Haobing Zhan and Dehao Zhang and Hao Zhang and Jin Zhang and Puqi Zhang and Qiao Zhang and Rui Zhang and Xiaobin Zhang and Xiaoyun Zhang and Y. Zhang and Yadong Zhang and Yangkun Zhang and Yichi Zhang and Yizhi Zhang and Yongting Zhang and Yu Zhang and Yushun Zhang and Yutao Zhang and Yutong Zhang and Zheng Zhang and Chenguang Zhao and Feifan Zhao and Jinxiang Zhao and Shuai Zhao and Xiangyu Zhao and Xuanle Zhao and Yikai Zhao and Zijia Zhao and Huabin Zheng and Ruihan Zheng and Shaojie Zheng and Tengyang Zheng and Junfeng Zhong and Longguang Zhong and Weiming Zhong and M. Zhou and Runjie Zhou and Xinyu Zhou and Zaida Zhou and Jinguo Zhu and Liya Zhu and Xinhao Zhu and Yuxuan Zhu and Zhen Zhu and Jingze Zhuang and Weiyu Zhuang and Ying Zou and Xinxing Zu},
  doi = {10.48550/arXiv.2602.02276},
  eprint = {2602.02276},
  primaryclass = {cs.CL},
  title = {{Kimi K2.5}: Visual Agentic Intelligence},
  url = {https://arxiv.org/abs/2602.02276},
  year = {2026}
}

@misc{deepseek2024v3,
  archiveprefix = {arXiv},
  author = {{DeepSeek-AI}},
  doi = {10.48550/arXiv.2412.19437},
  eprint = {2412.19437},
  primaryclass = {cs.CL},
  title = {{DeepSeek-V3} Technical Report},
  url = {https://arxiv.org/abs/2412.19437},
  year = {2024}
}

@misc{openai2025gptoss,
  archiveprefix = {arXiv},
  author = {{OpenAI}},
  doi = {10.48550/arXiv.2508.10925},
  eprint = {2508.10925},
  primaryclass = {cs.CL},
  title = {{gpt-oss-120b} \& {gpt-oss-20b} Model Card},
  url = {https://arxiv.org/abs/2508.10925},
  year = {2025}
}

@inproceedings{wang2023selfconsistency,
  author = {Xuezhi Wang and Jason Wei and Dale Schuurmans and Quoc V. Le and Ed H. Chi and Sharan Narang and Aakanksha Chowdhery and Denny Zhou},
  booktitle = {ICLR 2023},
  title = {Self-Consistency Improves Chain of Thought Reasoning in Language Models},
  url = {https://arxiv.org/abs/2203.11171},
  year = {2023}
}

@inproceedings{wang2025moa,
  author = {Wang, Junlin and Wang, Jue and Athiwaratkun, Ben and Zhang, Ce and Zou, James Y},
  booktitle = {International Conference on Learning Representations},
  editor = {Y. Yue and A. Garg and N. Peng and F. Sha and R. Yu},
  pages = {33944--33963},
  title = {Mixture-of-Agents Enhances Large Language Model Capabilities},
  url = {https://proceedings.iclr.cc/paper_files/paper/2025/file/5434be94e82c54327bb9dcaf7fca52b6-Paper-Conference.pdf},
  volume = {2025},
  year = {2025}
}

\appendix
\raggedbottom
\noindent The appendices describe the inference procedures
(Appendix~\ref{sec:appendix-methods}), analysis
(Appendix~\ref{sec:appendix-analysis}), preprocessing
(Appendix~\ref{sec:appendix-preprocess}), and evaluation
(Appendix~\ref{sec:appendix-metrics}). Pseudocode exposes the experimental
control flow while omitting API, logging, and \mbox{serialization boilerplate}.
Prompt descriptions below summarize the instructions rather than
reproduce their literal \mbox{API representation}.

\section{Inference Procedures}
\label{sec:appendix-methods}

The primary backbone is Qwen3.5-397B-A17B.
All conditions receive the same original image and benchmark-specific
query. Let $M$ denote the backbone, $H$ the interaction history, and
$\mathcal{T}_{\mathrm{ext}}$ the Search, Visit, and Python actions.
The configuration names \texttt{wotool}, \texttt{tool},
\texttt{withthink}, and \texttt{combine} correspond to
\emph{w/o Tools}, \emph{w/ Tools}, \textsc{ReVision}, and
Dual-Rollout Combine, respectively.

\subsection{Shared Inference Procedure}

Algorithm~\ref{alg:controller} is shared by the three single-trajectory
conditions. Tool availability is the configuration change:
$\mathcal{T}=\emptyset$ for \emph{w/o Tools},
$\mathcal{T}=\mathcal{T}_{\mathrm{ext}}$ for \emph{w/ Tools}, and
$\mathcal{T}=\mathcal{T}_{\mathrm{ext}}\cup\{\textsc{ReVision}\}$
for our method. The runner executes at most the first returned action
per turn, for at most 40 turns. Assistant reasoning is retained in the
record for analysis.

\begin{pseudocode}{Shared MLLM rollout}{alg:controller}
\pinput{Backbone $M$, initial history $H_0$, actions $\mathcal{T}$.}
\poutput{Final answer and recorded history $H$.}
\pstep{$H \leftarrow \operatorname{Copy}(H_0)$.}
\pstep{\textbf{for} $t=1,\ldots,40$ \textbf{do}}
\pstep[1]{$a \leftarrow M(H,\mathcal{T})$, keeping its first action call.}
\pstep[1]{Append $a$ and its reasoning to $H$.}
\pstep[1]{\textbf{if} $a$ has no action call \textbf{then}}
\pstep[2]{\textbf{return} parsed answer from $a$, and $H$.}
\pstep[1]{\textbf{if} the action is \textsc{ReVision} \textbf{then}}
\pstep[2]{$o \leftarrow \operatorname{Reconsider}(M,H,a.\mathrm{thought})$.}
\pstep[1]{\textbf{else} execute the external action to obtain $o$.}
\pstep[1]{Append $o$ as the matching tool result to $H$.}
\pstep{\textbf{return} parsed content of the last nonempty assistant
message, or the budget-exhaustion marker if none exists.}
\end{pseudocode}

\subsection{ReVision Interface and Policy}

The action \texttt{plan\_and\_reflect} takes one string argument,
\texttt{thought}. Its description invites the
MLLM to plan before searching, integrate collected evidence, and
decide whether to search further or answer. The argument contains the
MLLM's current knowledge, uncertainty, and intended \mbox{next step}.

The MLLM policy balances external fact gathering with intrinsic
reasoning and image analysis. It recommends planning before search,
reviewing evidence after two or three searches, and reconsidering the
image and question when search stalls. These are prompt instructions.
The runner does not schedule or force a reasoning action. The policy
also asks the MLLM to verify specialized facts through search
and give a final answer only when evidence is sufficient.

Benchmark-specific suffixes specify the answer format. EXAMS-V asks
the model to read the question and choices in the image and return an
option letter. VisualSimpleQA requests a short factual answer. CSVQA
requires the boxed final-answer format. The remaining suffixes ask for
the final option or answer once evidence is sufficient.

\subsection{Nested Reasoning Call}

Algorithm~\ref{alg:revision} excludes the unresolved action-call
message before copying the history. The original multimodal user
message, all completed assistant turns, and external observations
remain available. The isolated call uses the same backbone with all
tools disabled.

\begin{pseudocode}{Image-grounded reconsideration}{alg:revision}
\pinput{$M$, history $H$ ending in a pending call, thought $u$.}
\poutput{A tool observation $o$ for the MLLM.}
\pstep{$H' \leftarrow \operatorname{Copy}(H\text{ without its last message})$.}
\pstep{Append $u$ and the pure-reasoning instruction to $H'$.}
\pstep{Call $b \leftarrow M(H',\emptyset)$.}
\pstep{\textbf{if} the call fails \textbf{then}}
\pstep[1]{\textbf{return} an observation retaining $u$ and the error.}
\pstep{\textbf{if} $b$ attempts a tool call \textbf{then}}
\pstep[1]{$d \leftarrow$ reminder that tools are disabled.}
\pstep{\textbf{else} $d \leftarrow$ first nonempty value among
$b$'s content, reasoning, and a continuation reminder.}
\pstep{Save $b$'s reasoning separately for analysis.}
\pstep{\textbf{return} an observation containing $u$ and $d$.}
\end{pseudocode}

The pure-reasoning instruction presents the MLLM's thought,
asks for deeper reasoning about the next step, and explicitly forbids
tool use. It directs the model to defer any further search or final
answer to the next MLLM turn. The returned analysis is inserted
as the result of the original action. The MLLM then resumes
with all four \mbox{actions available}.

\subsection{Think-Scratchpad Control}

The scratchpad ablation retains the same interface but disables the
extra model call. Its observation simply echoes the MLLM's
\texttt{thought}, making it visible in the next turn. It performs no
retrieval or nested reasoning. The registered-stub ablation instead
returns a placeholder, while keeping the interface and MLLM
prompt fixed.

\subsection{Dual-Rollout Combine}

Combine independently generates one \emph{w/o Tools} trajectory and
one \emph{w/ Tools} trajectory per task. The fusion input keeps
assistant reasoning, textual responses, action calls, textual tool
observations, and final answers. System messages and repeated user
messages are omitted. Assistant responses and tool observations are
capped at 1,500 and 4,000 characters, respectively. Image payloads in
tool observations are excluded. The original image is \mbox{supplied again}.

\begin{pseudocode}{Dual-Rollout Combine}{alg:combine}
\pinput{$M$, original image $I$, query $q$, answer-format rule $f$.}
\poutput{Fused final answer.}
\pstep{$H_0 \leftarrow$ initial messages containing $I$ and $q$.}
\pstep{$(y_0,H_{\mathrm{wo}}) \leftarrow \operatorname{Rollout}(M,H_0,\emptyset)$.}
\pstep{$(y_1,H_{\mathrm{wt}}) \leftarrow \operatorname{Rollout}(M,H_0,\mathcal{T}_{\mathrm{ext}})$.}
\pstep{Serialize both histories with the retention rules above.}
\pstep{Construct the fusion query from $q$, both serialized
histories, their final answers, and $f$.}
\pstep{Call $M$ with the synthesis instruction, fusion query,
and original image $I$, with all tools disabled.}
\pstep{\textbf{return} the parsed final answer.}
\end{pseudocode}

The synthesis instruction asks the model to compare the quality and
thoroughness of both approaches, identify errors and inconsistencies,
assess their suitability for the question, and synthesize the most
reliable answer. It explicitly cautions that tool-based reasoning can
contain implementation errors and reasoning without tools can contain
calculation mistakes. A benchmark-specific answer-format suffix is
appended to the user query.

\section{Accuracy and Reasoning-Token Analysis}
\label{sec:appendix-analysis}

\subsection{Accuracy Aggregation}

For a run with $N$ evaluated tasks and binary scores $s_i$, accuracy
is $100N^{-1}\sum_i s_i$. The primary backbone uses three runs:
we report the mean of their accuracies and the sample standard
deviation. For benchmark accuracies $A_b$ and sizes $N_b$,
\begin{align}
 A_{\mathrm{macro}} &= \frac{1}{B}\sum_{b=1}^{B} A_b,\\
 A_{\mathrm{micro}} &= \frac{\sum_b N_b A_b}{\sum_b N_b}.
\end{align}
The Kimi-K2.5 transfer experiment uses one run per condition.

\subsection{Paired Answer Reversals}

Paired comparisons join records by \texttt{task\_id} and exclude
items missing from either condition. Algorithm~\ref{alg:pairing}
constructs the four correctness cells. The descriptive reversal
rate uses all common tasks as its denominator. A conditional loss
rate among previously solved tasks instead divides by the number
correct without tools.

\begin{pseudocode}{Paired correctness analysis}{alg:pairing}
\pinput{Correctness maps $s_{\mathrm{wo}}$ and $s_{\mathrm{wt}}$.}
\poutput{Four cell counts and paired reversal rate.}
\pstep{$J \leftarrow \operatorname{IDs}(s_{\mathrm{wo}})\cap
\operatorname{IDs}(s_{\mathrm{wt}})$.}
\pstep{Initialize $C_{00},C_{01},C_{10},C_{11}\leftarrow 0$.}
\pstep{\textbf{for each} $j\in J$ \textbf{do}}
\pstep[1]{$u\leftarrow s_{\mathrm{wo}}(j)$, $v\leftarrow s_{\mathrm{wt}}(j)$.}
\pstep[1]{$C_{uv}\leftarrow C_{uv}+1$.}
\pstep{$r\leftarrow C_{10}/|J|$ if $J\ne\emptyset$, otherwise $0$.}
\pstep{\textbf{return} the cell counts, $|J|$, and $r$.}
\end{pseudocode}

\subsection{Reasoning-Token Accounting}

Let $\ell(x)$ be the token count under the backbone's official
tokenizer, and let $r_t$ be the nonempty assistant reasoning in
MLLM turn $t$. The full-trajectory MLLM count is
$R(H)=\sum_t\ell(r_t)$. For an executed \textsc{ReVision} action
$k$, let $u_k$ be its argument and $d_k$ its returned analysis.
Writing RV for ReVision and C for Combine, the per-query totals are
\begin{align}
 R_{\mathrm{RV}} &= R(H)+\sum_k[\ell(u_k)+\ell(d_k)],\\
 R_{\mathrm{C}} &= R(H_{\mathrm{wo}})+R(H_{\mathrm{wt}})+R(H_{\mathrm{fuse}}).
\end{align}
These definitions count the MLLM argument and returned
\texttt{deep\_thought} for ReVision, and both source trajectories
plus the fusion call for Combine. The saved nested reasoning field
is not added again as a separate MLLM turn.

\begin{pseudocode}{MLLM reasoning summaries}{alg:tokens}
\pinput{Query records $\mathcal{Q}$ and official tokenizer.}
\poutput{Mean total tokens and mean tokens per reasoning turn.}
\pstep{\textbf{for each} query $q\in\mathcal{Q}$ \textbf{do}}
\pstep[1]{$L_q\leftarrow[\ell(r_t):r_t\text{ is nonempty assistant reasoning}]$.}
\pstep[1]{$T_q\leftarrow\sum L_q$.}
\pstep[1]{$U_q\leftarrow\operatorname{Mean}(L_q)$ if nonempty, otherwise $0$.}
\pstep{\textbf{return} $\operatorname{Mean}_q T_q$,
$\operatorname{Mean}_q U_q$, and the count of queries with $T_q=0$.}
\end{pseudocode}

The turn-level summary is computed per query before averaging
across queries. It is therefore distinct from pooling all reasoning
turns into a single benchmark-level average.

\section{Dataset Preprocessing}
\label{sec:appendix-preprocess}

\subsection{Sources and Output Contract}

Table~\ref{tab:appendix-splits} lists the evaluated splits.
Official reader outputs are adapted to one JSONL record per task.
The record contains \texttt{key\_question}, \texttt{key\_image},
\texttt{task}, \texttt{prompt}, \texttt{gt}, \texttt{metadata},
\texttt{image}, and \texttt{image\_byte}. The ground-truth field
stores the answer, question type, answer type, precision, and choices
as JSON. Metadata retains the source identifier, benchmark fields,
and relative and absolute image paths.

\begin{appendixtable}
\centering
\small
\setlength{\tabcolsep}{3pt}
\begin{tabular}{@{}lrr@{}}
\toprule
Benchmark & Split & Tasks \\
\midrule
CSVQA & train & 1,378 \\
EXAMS-V & test & 3,565 \\
MSEarth-MCQ & train & 2,785 \\
VisualSimpleQA & test & 500 \\
WikiVQABench & train & 344 \\
WorldVQA & train & 3,000 \\
\midrule
Total & & 11,572 \\
\bottomrule
\end{tabular}
\caption{Official release splits used for evaluation. A release
split named ``train'' is used as a test benchmark. No model training
is performed.}
\label{tab:appendix-splits}
\end{appendixtable}

Images are decoded to RGB, saved as JPEG at quality 95, and encoded
in base64 for \texttt{image\_byte}. The source image content is used
without OCR or question regeneration. In particular, EXAMS-V retains
the question, options, and diagrams together in its image.

\subsection{Benchmark-Specific Adapters}

\paragraph{CSVQA.}
The official TSV reader selects language-specific question and option
columns, removes the image marker, and retains nonmissing options.
The official prompt includes the question category and boxed
final-answer instruction.

\paragraph{EXAMS-V.}
The adapter uses \texttt{sample\_id} and \texttt{answer\_key}.
The user instruction asks the model to read the image and return
one option letter from A--E. No question text is reconstructed.

\paragraph{MSEarth-MCQ.}
The released \texttt{query}, containing caption, question, and
options, is copied directly. The answer is the leading option
identifier in \texttt{response}, before the first period.

\paragraph{VisualSimpleQA.}
The adapter uses \texttt{multimodal\_question} and the released
answer. It retains the text-only question, evidence, category,
and rationale in metadata.

\paragraph{WikiVQABench.}
The displayed choices are the correct answer followed by the
released distractors. Options are lettered in that order, so the
stored ground-truth option is A. The correct answer text and
complete choice list remain in metadata.

\paragraph{WorldVQA.}
The released question is prefixed with the language-specific
instruction to provide as much detail as possible. The source
index, language, category, and difficulty are retained.

\begin{pseudocode}{Benchmark-to-JSONL adaptation}{alg:preprocessing}
\pinput{Benchmark $b$, official reader, output directory.}
\poutput{One JSONL record per evaluated task.}
\pstep{Load the release and split in Table~\ref{tab:appendix-splits}.}
\pstep{\textbf{for each} source row $x$ \textbf{do}}
\pstep[1]{Construct its query, answer, choices, and metadata
with the benchmark-specific adapter above.}
\pstep[1]{Decode and save the image, then encode the saved bytes.}
\pstep[1]{Create the standard record with the source task ID.}
\pstep{Verify the task count against Table~\ref{tab:appendix-splits}.}
\pstep{Write records as UTF-8 JSONL, one record per line.}
\end{pseudocode}

\subsection{Data Provenance}

\begingroup\raggedright
The release identifiers are \path{Skywork/CSVQA},
\path{MBZUAI/EXAMS-V}, \path{PrismaX/MSEarth_MCQ},
\path{WYLing/VisualSimpleQA} (configuration
\texttt{VisualSimpleQA}), \path{ibm-research/WikiVQABench},
and \path{moonshotai/WorldVQA}.
The corresponding official resources are
\href{https://github.com/csvqa-benchmark/CSVQA}{CSVQA},
\href{https://github.com/mbzuai-nlp/EXAMS-V}{EXAMS-V},
\href{https://github.com/xiangyu-mm/MSEarth}{MSEarth},
\href{https://huggingface.co/datasets/WYLing/VisualSimpleQA}{VisualSimpleQA},
\href{https://github.com/basels/VLMEvalKit/tree/wiki-vqa-bench}{WikiVQABench},
\mbox{and \href{https://github.com/MoonshotAI/WorldVQA}{WorldVQA}}.
\par\endgroup

\section{Evaluation Protocols}
\label{sec:appendix-metrics}

\subsection{Multiple-Choice Matching}

The MCQ benchmarks are CSVQA, EXAMS-V, MSEarth-MCQ, and
WikiVQABench. Algorithm~\ref{alg:evaluation} summarizes the parser
and the open-ended evaluation path. In the implementation,
the boxed-option pattern uses the first regex match. If it is
absent, the parser uses the last standalone \mbox{A--E letter}.

\begin{pseudocode}{Per-task correctness evaluation}{alg:evaluation}
\pinput{Benchmark $b$, question $q$, reference $g$, prediction $y$.}
\poutput{Binary correctness, plus category for open-ended tasks.}
\pstep{\textbf{if} $b$ is an MCQ benchmark \textbf{then}}
\pstep[1]{$p\leftarrow$ first boxed A--E option in $y$, if present.}
\pstep[1]{Otherwise use the last standalone A--E letter in $y$.}
\pstep[1]{Normalize the reference to its first option letter.}
\pstep[1]{\textbf{return} $\mathbf{1}[p\ne\emptyset\ \land\ p=g]$.}
\pstep{\textbf{else} select $b$'s official judge and prompt.}
\pstep[1]{Evaluate $(q,g,y)$ at temperature $0$.}
\pstep[1]{Parse the judge's label and retain its raw output.}
\pstep[1]{\textbf{return} $1$ only for the correct category,
otherwise $0$, together with the category.}
\end{pseudocode}

\subsection{Open-Ended Judging}

VisualSimpleQA uses DeepSeek-V3~\citep{deepseek2024v3}. WorldVQA uses
\texttt{gpt-oss-120b}~\citep{openai2025gptoss}. Both judges run at temperature 0 with
the respective benchmark's official prompt. The protocols retain
distinct label semantics rather than substitute one generic judge.
An unparseable judge label is an evaluation error, not a wrong answer.

\paragraph{VisualSimpleQA.}
The judge receives the question, reference (Answer\_1), and
prediction (Answer\_2). Label $1$ requires all reference information
without additional incorrect information. Semantically equivalent
wording is accepted. Label $0$ denotes explicit refusal, and $-1$
denotes an incorrect attempted answer or missing reference information.
The prompt requests only the label. The parser extracts a standalone
$-1$, $0$, or $1$ and saves the category and raw response.

\paragraph{WorldVQA.}
The judge ignores formatting, punctuation, language, and
abbreviation differences while assessing semantics and granularity.
A correct answer contains the reference's core information without
contradiction and is equally or more specific. A coarser answer,
wrong entity, or wrong value is incorrect. Refusal, referral to
external search, or an answer with neither relevant information
nor contradiction is unattempted. The output contains an
\emph{Evaluation} explanation followed by a \emph{Label}. Only
\texttt{Correct} earns one point.

\subsection{WorldVQA Few-Shot Examples}

The official judge prompt includes six cases.
Table~\ref{tab:judge-examples} summarizes their decisions.
Chinese examples are described in English here for readability.
These descriptions explain the rubric and do not replace the
examples used by the evaluator.

\begin{appendixtable}
\centering
\small
\setlength{\tabcolsep}{3pt}
\begin{tabular}{@{}p{0.69\columnwidth}p{0.25\columnwidth}@{}}
\toprule
Prediction relative to reference & Label \\
\midrule
Generic terraces instead of Longji terraces & Incorrect \\[3pt]
A severe storm, tornado, or hurricane for ``high winds'' & Correct \\[3pt]
``via browser'' for the brand ``via'' & Correct \\[3pt]
Cannot identify the athlete Wout Weghorst & Unattempted \\[3pt]
An uncertain guess of Aksu apples instead of Yantai apples & Incorrect \\[3pt]
Suggests expert identification instead of naming the Japanese rhinoceros beetle & Unattempted \\
\bottomrule
\end{tabular}
\caption{Summaries of the six official WorldVQA few-shot cases.
They illustrate granularity, semantic agreement, incorrect guesses,
and unattempted answers.}
\label{tab:judge-examples}
\end{appendixtable}

\end{document}